\documentclass[letterpaper, 10 pt, conference]{ieeeconf}  

\IEEEoverridecommandlockouts                              

\usepackage{graphics} 
\usepackage{epsfig} 
\usepackage{mathptmx} 
\usepackage{times} 
\usepackage{amsmath} 
\usepackage{amssymb}  
\let\labelindent\relax
\usepackage{enumitem}
\usepackage{subfig}
\usepackage{graphicx}
\usepackage[symbol]{footmisc}

\usepackage{url}

\title{\LARGE \bf
Virtual Reality Simulator for Fetoscopic Spina Bifida Repair Surgery
}

\author{Przemysław Korzeniowski$^{1}$, Szymon Płotka$^{1, 2}$, Robert Brawura-Biskupski-Samaha$^{3}$, Arkadiusz Sitek$^{1}$
\thanks{$^{1}$Przemysław Korzeniowski, Szymon Płotka and Arkadiusz Sitek are with Sano Centre for Computational Medicine, Czarnowiejska 36, Cracow, Poland
        {\tt\small \{p.korzeniowski, s.plotka, a.sitek\}@sanoscience.org}}%
\thanks{$^{2}$ Szymon Płotka is with Quantitative Healthcare Analysis (qurAI) group, Informatics Institute, University of Amsterdam, Science Park 904, Amsterdam, The Netherlands
        {\tt\small s.s.plotka@uva.nl}}%
\thanks{$^{3}$ Robert Brawura-Biskupski-Samaha is with II Department of Obstetrics and Gynaecology, The Medical Centre of Postgraduate Education, Marymoncka 99/103, Warsaw, Poland 
        {\tt\small robertsamaha@gmail.com}}%
}

\begin{document}

\maketitle
\thispagestyle{empty}
\pagestyle{empty}

\begin{abstract}

Spina Bifida (SB) is a birth defect developed during the early stage of pregnancy in which there is incomplete closing of the spine around the spinal cord. The growing interest in fetoscopic Spina-Bifida repair, which is performed in fetuses who are still in the pregnant uterus, prompts the need for appropriate training. The learning curve for such procedures is steep and requires excellent procedural skills. Computer-based virtual reality (VR) simulation systems offer a safe, cost-effective, and configurable training environment free from ethical and patient safety issues. However, to the best of our knowledge, there are currently no commercial or experimental VR training simulation systems available for fetoscopic SB-repair procedures.
In this paper, we propose a novel VR simulator for core manual skills training for SB-repair.
An initial simulation realism validation study was carried out by obtaining subjective feedback (face and content validity) from 14 clinicians. The overall simulation realism was on average marked 4.07 on a 5-point Likert scale (1 - ‘very unrealistic’, 5 - ‘very realistic’). 
Its usefulness as a training tool for SB-repair as well as in learning fundamental laparoscopic skills was marked 4.63 and 4.80, respectively. 
These results indicate that VR simulation of fetoscopic procedures may contribute to surgical training without putting fetuses and their mothers at risk. It could also facilitate wider adaptation of fetoscopic procedures in place of much more invasive open fetal surgeries.


\end{abstract}

\section{Introduction}

Spina Bifida (SB) is the most common congenital birth defect affecting the central nervous system \cite{detrait2005human}. It usually manifests with sacs covered with skin in the lumbar region of the fetus back. However, in the most severe form of SB a section of the spinal cord can protrude through the back of the baby so that the spinal cord is fully exposed. This causes serious neurological disabilities after birth, which are related to the location and severity of the defect. The prevalence of SB is about 3.4 per 10000 live births in the USA \cite{boulet2008trends}. Newborns with untreated SB can have permanent nerve damage and live with major disabilities, including paralysis and bowel and bladder dysfunction \cite{mitchell2004spina}. About 10\% of newborns with a severe form of SB die \cite{manning2000pathophysiology}.

\begin{figure*}[t!]
    \centering
    \vspace*{0.2cm}
    \subfloat[]{\includegraphics[width=4.2cm, height=3.8cm]{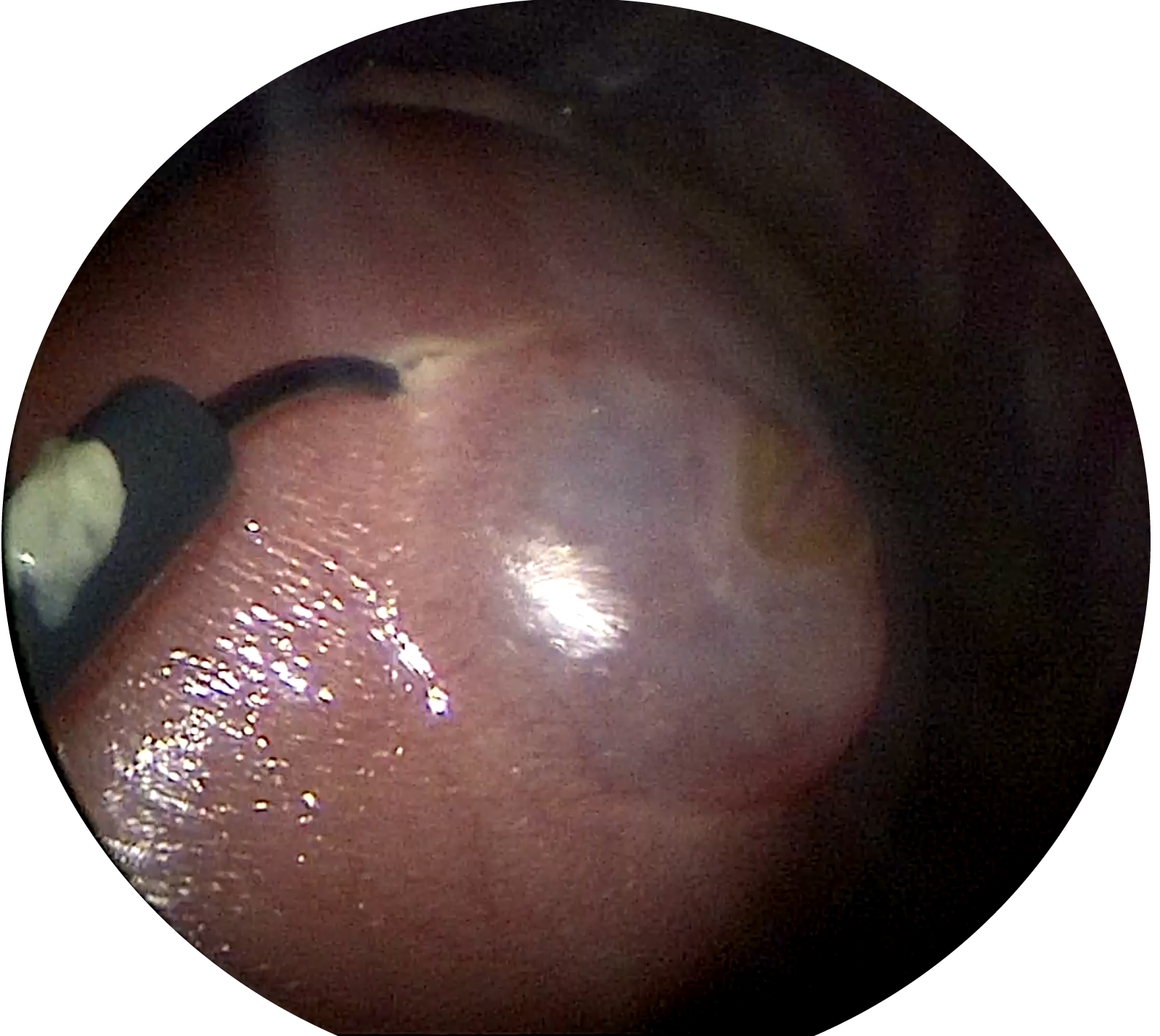}} 
    \subfloat[]{\includegraphics[width=4.2cm, height=3.8cm]{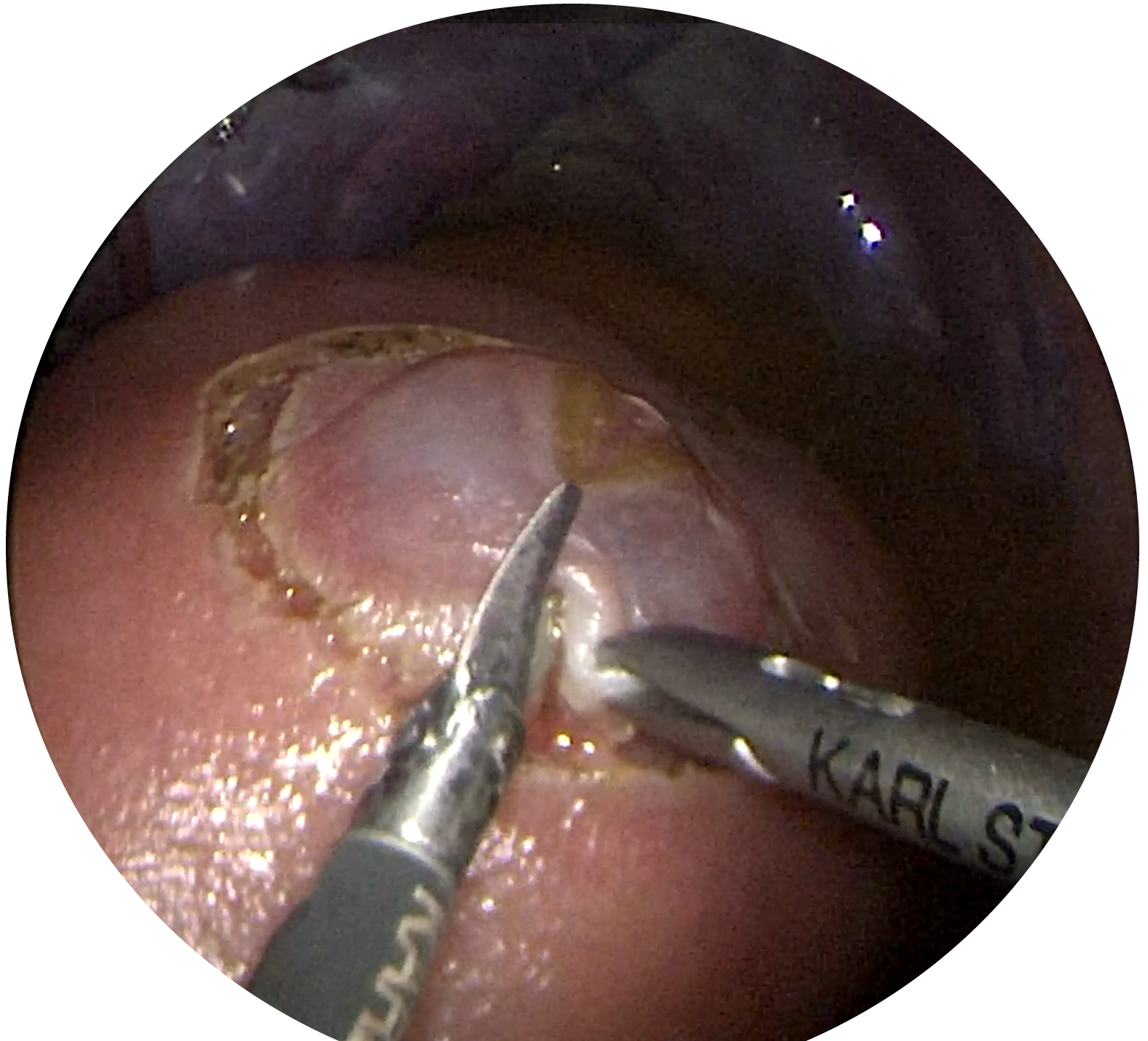}} 
    \subfloat[]{\includegraphics[width=4.2cm, height=3.8cm]{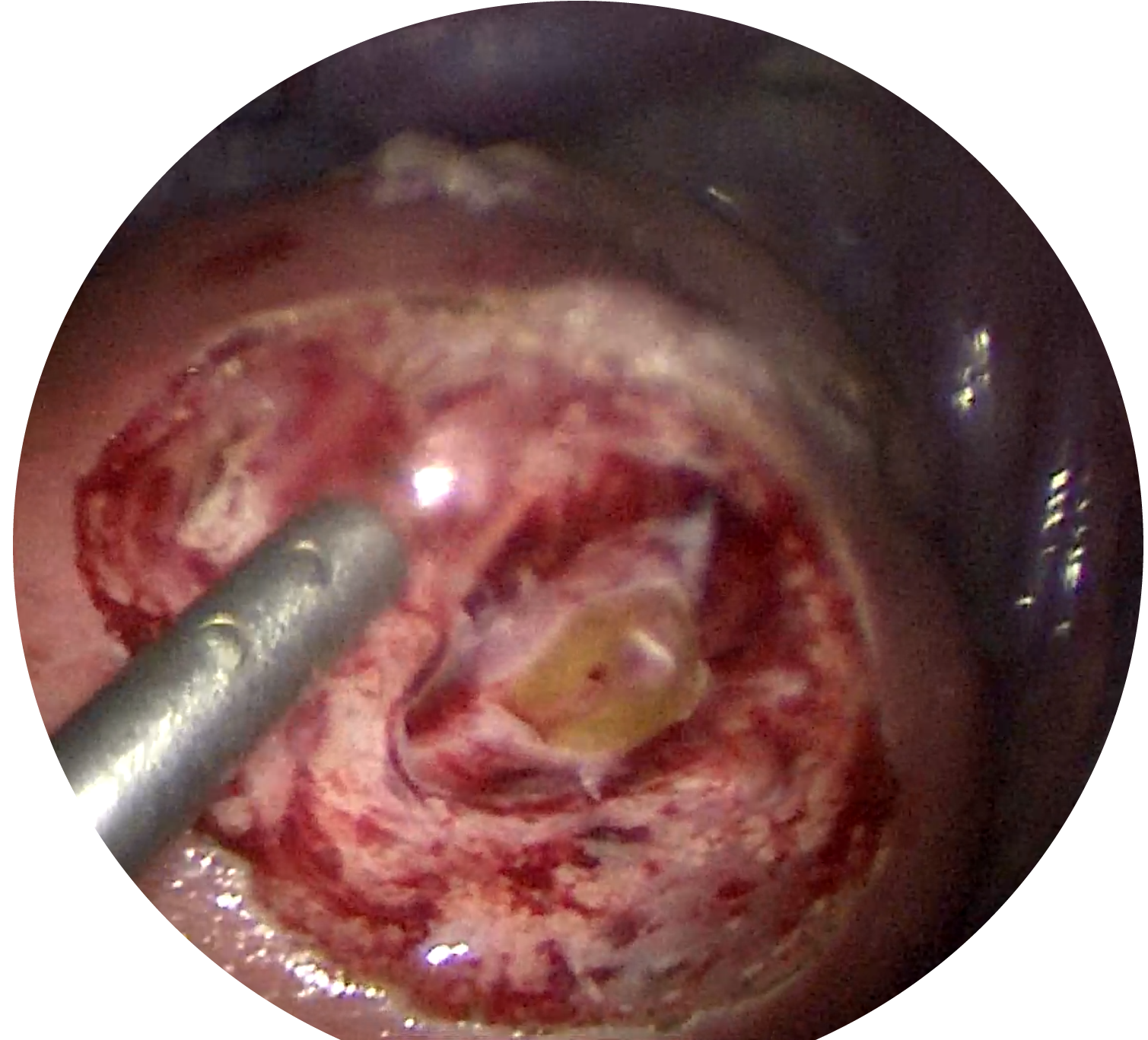}}
    \subfloat[]{\includegraphics[width=4.2cm, height=3.8cm]{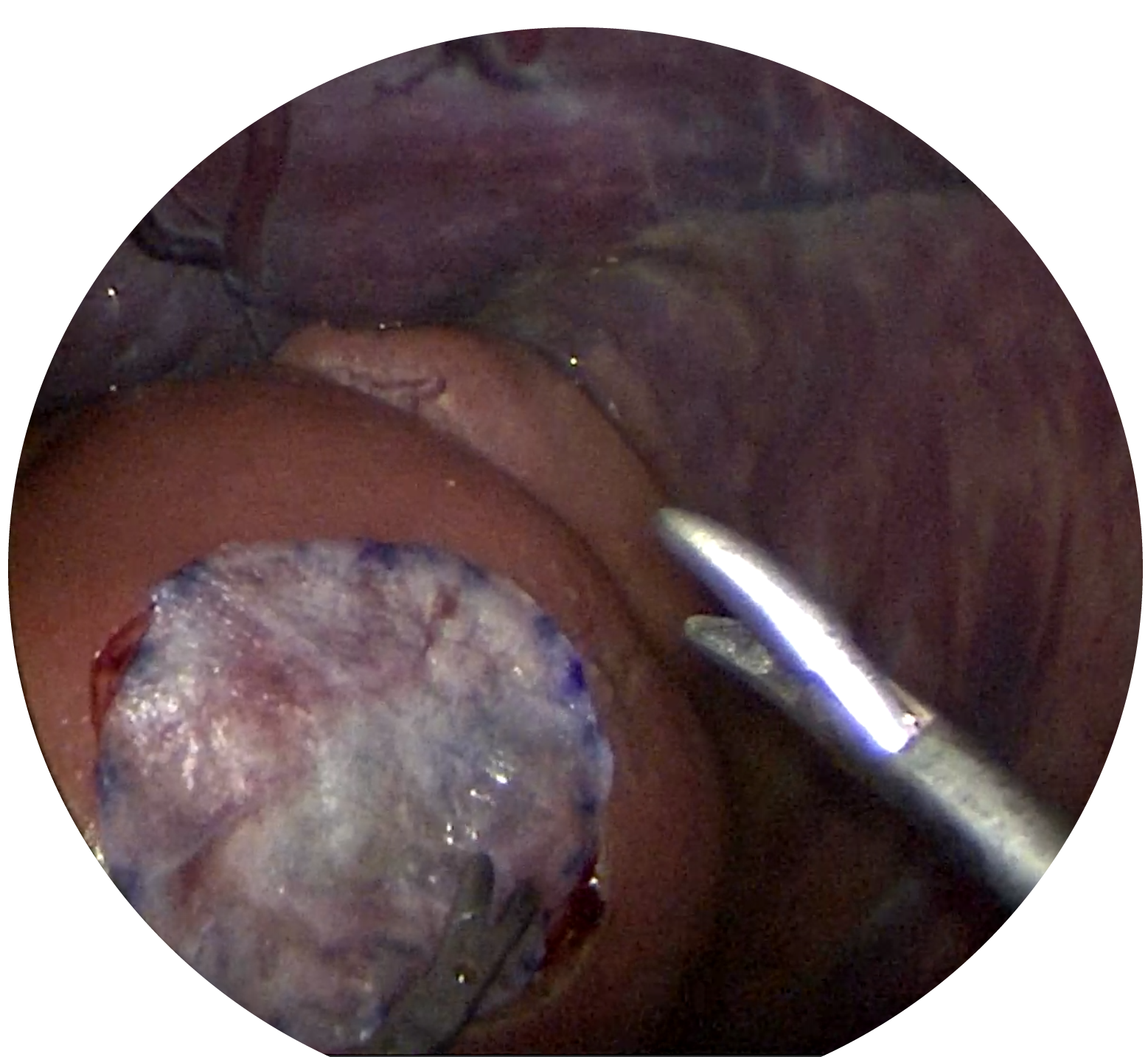}}
    \caption{Fetoscope images taken at various stages of SB repair. From the left diathermy, cutting, irrigation, and patching.}
    \label{fig:sb_foobar}
\end{figure*}

Prenatal repair of SB, which is performed in fetuses who are still in the pregnant uterus, has shown better outcomes compared to treatment of the condition after birth \cite{adzick2011randomized}. Although the efficacy of prenatal SB repair is well-proven, it is a complex and difficult medical intervention consisting of multiple stages.
During the first stage, a monopolar thermocoagulation (i.e. diathermy) is applied to make a circumferential incision around the placode (Fig.~\ref{fig:sb_foobar}~left). Next, the surrounding tissue is resected using a grasper and scissors (Fig.~\ref{fig:sb_foobar}~middle left) and the skin is undermined and cleaned using irrigation (Fig.~\ref{fig:sb_foobar}~middle right). Finally, myofascial flaps are sutured with a patch (Fig.~\ref{fig:sb_foobar}~right). 
For more details regarding the SB repair procedure we refer the reader to \cite{heuer2015fetal} and \cite{lapa2018percutaneous}. 

In addition to the complexity of the repair, the condition is relatively rare. Hence, there is a limited number of surgeons who are capable of performing such prenatal procedures. As a result, access to medical care for those affected with SB is limited. To improve this, increasing the training level of the surgeons is important. In this paper, we propose a prototype VR simulator for core manual skills training of fetoscopic SB-repair, which we developed in close collaboration with an expert fetoscopic surgeon.

\footnotetext[1]{\url{https://en.wikipedia.org/wiki/Spina_bifida}}

\begin{figure}[t!]
    \centering
    \includegraphics[width=8.5cm]{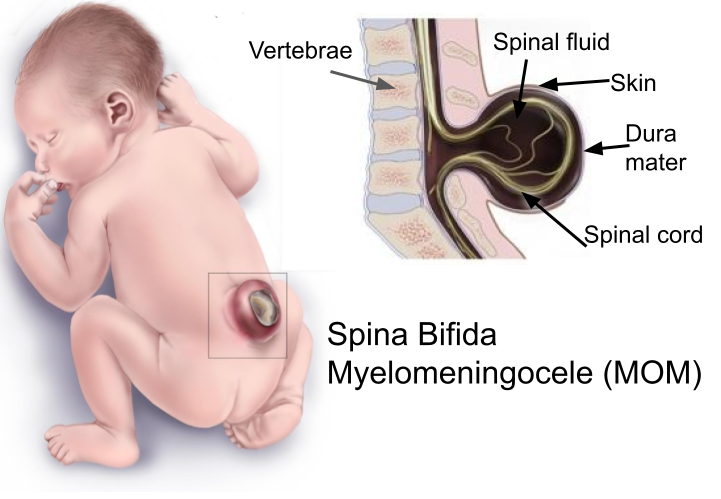}
    \caption{Depiction of SB. The top of SB is covered only with dura mater for the most severe type of SB myelomeningocele. Figure adopted from \protect\footnotemark[1].}
    \label{fig:my_label}
\end{figure}

Multiple specialities in surgery have found  minimally invasive physical trainers alongside virtual reality (VR) simulators useful for the improvement of surgical skills and confidence. As early as 2002 in a seminal work by Seymour {\em et al.} \cite{seymour2002virtual} the VR simulator was shown to significantly improve the performance of residents during laparoscopic cholecystectomy. In more recent work, Yiannakopoulou \cite{yiannakopoulou2015virtual} reported that VR simulators have similar efficiency in improving surgical skills to physical task trainers. However, task trainers lack the real-life effect of a complete surgery as well as physiological behaviour. Although there is undisputed potential value in VR simulators there is a need for thorough investigations that ideally use a series of realism validation trials of such systems \cite{moglia2016systematic} that confirms their usefulness in training curricula. 

Most of the work on VR simulators is focused on the adult population. The usefulness of VR technology in pediatric surgery was demonstrated by Korzeniowski {\em et al.} \cite{korzeniowski2021virtual} who recently presented a validated VR simulator for laparoscopic paediatric inguinal hernia repair. The development of a physical trainer for SB-repair has been previously reported in the literature. Most recently, Joyeux {\em et al.} \cite{joyeux2021validation} used the animal model to simulate the BS repair. However, to the best of our knowledge, the system presented here is the first VR simulator for fetoscopic SB repair. In this article, besides the simulation system description we also report the results of the initial realism evaluation. Our main contributions are:

\begin{enumerate}
    \item Development of the first VR simulation system for SB repair.
    \item Establishing the face and content validity of the developed simulator.
\end{enumerate}

The rest of the paper is organized as follows. In Section \ref{sec:methods}, we describe real-time physics-based simulation software components and algorithms used to develop virtual SB surgery. In Section \ref{sec:experiments} we describe our validation protocol and provide results of the simulation realism validation study. Section \ref{sec:conclusions}  discusses the research carried out, identifies limitations of the approaches used, and outlines possible future work.

\section{Methods}
\label{sec:methods}

\begin{figure}
    \centering
    \includegraphics[width=8.5cm]{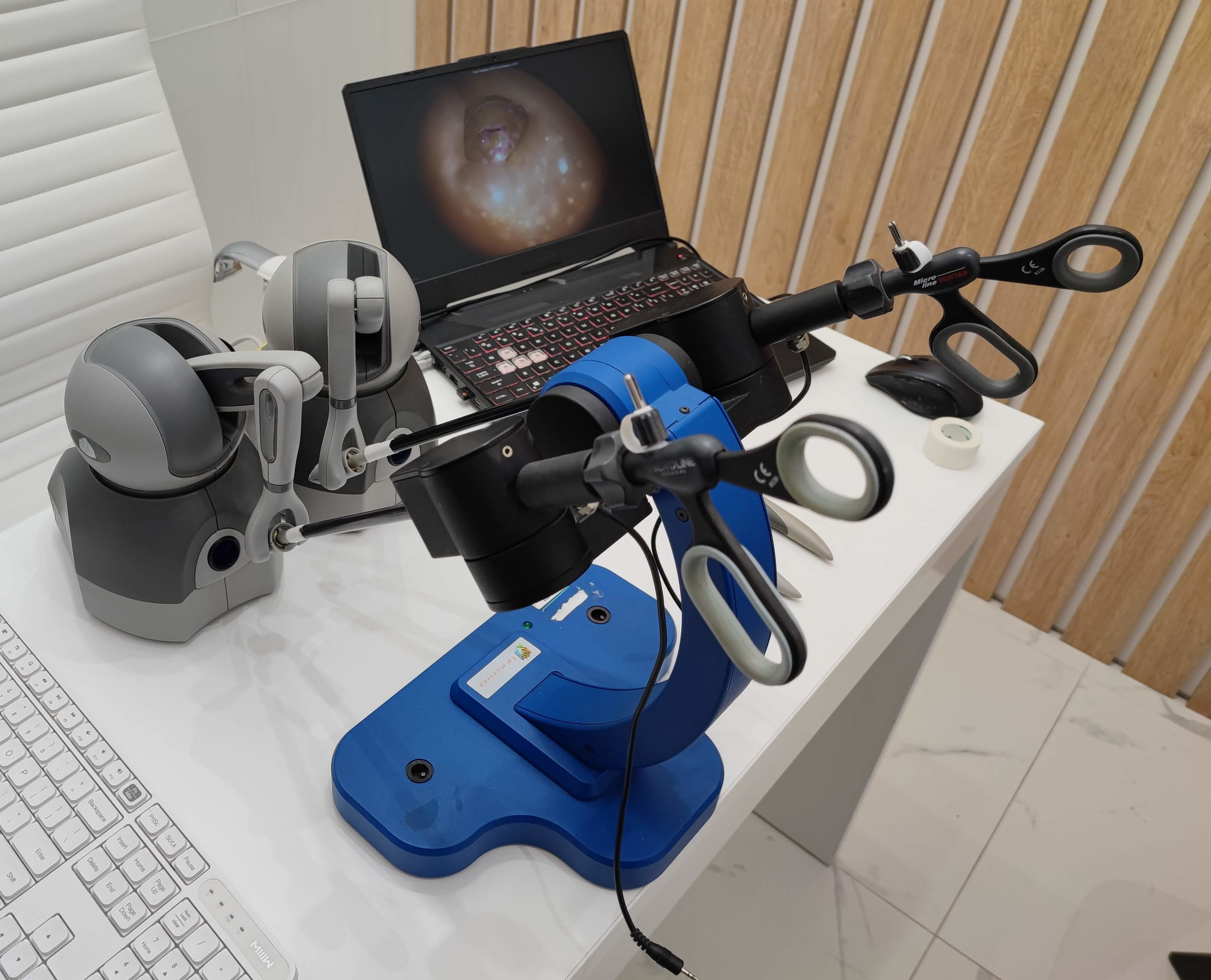}
    \caption{The overview of our setup. The hardware setup of the VR simulator for fetoscopic SB-repair consist of laptop, two Geomagic Touch haptic devices linked with Virtual Laparoscopic Interface.}
    \label{fig:hwsetup}
\end{figure}

\subsection{Description of Virtual Procedure}
\begin{figure*}[t!]
    \centering
    \vspace*{0.2cm}
    \subfloat[]{\includegraphics[width=5cm, height=4.5cm]{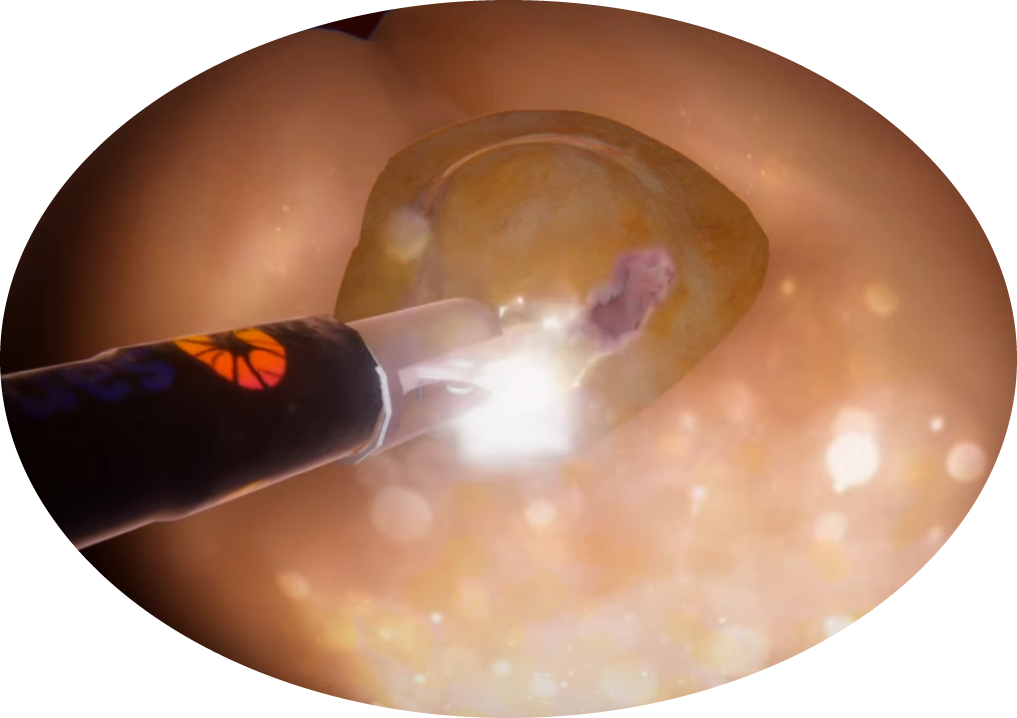}} \hspace{0.5cm} 
    \subfloat[]{\includegraphics[width=5cm, height=4.5cm]{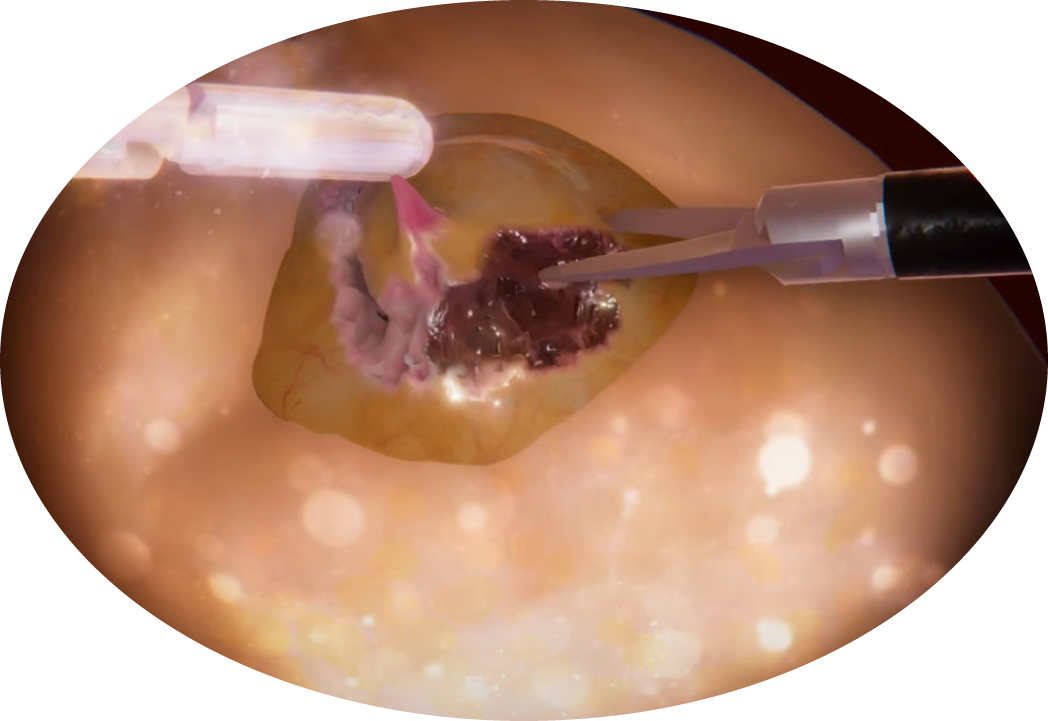}} \hspace{0.5cm}
    \subfloat[]{\includegraphics[width=5cm, height=4.5cm]{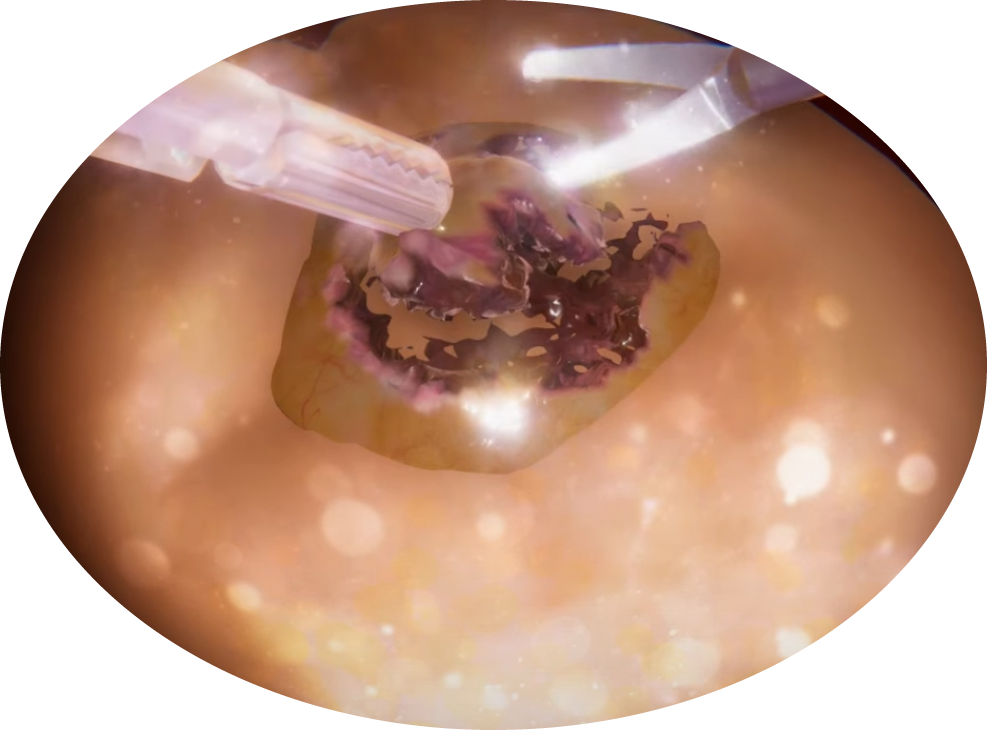}}
    \caption{Placode dissection with monopolar diathermy (left), tissue resection with scissors and grasper (center and right).}
    \label{fig:vr_surgery}
\end{figure*}

The virtual procedure starts with laparoscopic instruments inserted into the uterus inflated with carbon dioxide gas. The simulation visualizes the exposed lumbar region of the fetus back. The first stage of the simulation is to use a monopolar thermocoagulation (i.e. diathermy) to make a circumferential incision around the virtual placode (Fig.~\ref{fig:vr_surgery}~left). Next, the surrounding tissue is resected using virtual grasper and scissors (Fig.~\ref{fig:vr_surgery}~center and Fig.~\ref{fig:vr_surgery}~right). At this development stage, the simulator does not provide irrigation and suturing of myofascial flaps with a patch. In this paper, we solely focus on the realistic implementation of tissue dissection and resection using diathermy and scissors tools. Due to a small laparoscopic work-space and complex non-linear behaviour of thin tissues, realistic simulation of such procedure, especially in real-time at haptic interactive rates, is challenging.

\subsection{Simulation System Overview}

The experimental apparatus consisted of real-time software running on a laptop PC (Ryzen 7 5800H@3.2 GHz CPU, 32GB RAM and NVIDIA RTX 3060 GPU). Two Geomagic Touch input devices (3DSystems, US) combined with Virtual Laparoscopic Interface (VLI, Immersion, US) to mimic the laparoscopic instrumentation were used. The simulation setup is presented in Fig.~\ref{fig:hwsetup}.

The Unity3D engine (Unity Technologies, San Francisco, USA) was used as a development platform in combination with an in-house medical simulation framework. A multi-threaded Gauss-Seidel \cite{XPBD} physics solver was implemented in C/C++. Despite the high computational demands refresh rates of up to 1.5 kHz were obtained and allowed for small simulation time steps (0.75ms – 1.0ms), which are essential for accurate simulation at the haptic interactive rates.

\subsection{Simulation Framework}

To tackle the complexity of the simulation of the non-linear behaviour of tissues and the simulation of the haptic response, we used the Position-Based Dynamics (PBD) \cite{PBD} as our simulation method. Due to the local nature of the non-linear Gauss-Seidel process, PBD avoids many pitfalls of global, matrix-based solvers, allowing for stable handling of equality and inequality constraints. 
Moreover, several more recent papers turned PBD into a serious competitor of more advanced methods in terms of accuracy, stability, speed, and simplicity. Following the work of \cite{XPBD} we used Extended PBD (XPBD), which addressed PBD’s long-standing problem of stiffness constraint dependence on the number of solver iterations and the size of the time step. This relatively minor modification allowed for an accurate and efficient simulation of arbitrary elastic and dissipative energy potentials in an implicit manner. In addition, XPBD provides constraint force estimates important in surgical simulations, which require accurate haptic force-feedback calculations. 

We split every time step into $n$ equal time sub-steps and perform a single constraint iteration of XPBD in each sub-step. This is following the work of \cite{SmallSteps} who observed that performing a single large time-step with $n$ constraint solver iterations is less effective than computing $n$ smaller time-steps, each with a single constraint solver iteration. This approach was further validated in \cite{RigidBodies} with authors implementing a fully-fledged rigid body solver that handles surface contacts, a variety of joint types and the interaction with soft objects in a unified way. This is particularly interesting in the context of surgical robotics, as it allows for straightforward interactions of robots' rigid surgical tools with soft-tissues.

In \cite{XPBD} and \cite{SmallSteps} the authors compared XPBD with sub-stepping to more sophisticated solvers, both implicit or explicit, and stated that it produces visually similar results. This was achieved while maintaining the implementation straightforwardness of the original PBD method and being non-sensitive to matrix ill-conditioning.

\subsection{Implementation Details}

\subsubsection{Soft tissue modelling}

\begin{figure}
    \centering
    \includegraphics[width=8.5cm]{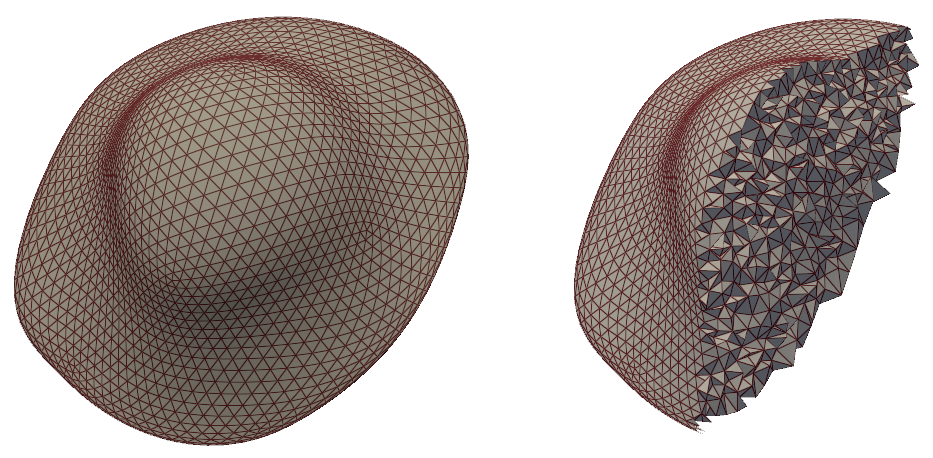}
    \caption{On the left: Volume discretization using tetrahedral mesh. On the right: Cut-through the tetrahedral mesh}
    \label{fig:tetras}
\end{figure}

The virtual anatomy of SB tissue was modelled as a volumetric tetrahedral mesh consisting of approximately 40k elements (Fig.~\ref{fig:tetras}). As soft-tissues are nearly incompressible, we used Neo-Hookean constitutive model proposed in \cite{macklin2021constraint} as it provides better volume conservation than co-rotational finite element  or St.~Venant-Kirchoff models. Moreover, the Neo-Hookean model recovers from inverted states ({\em i.e.} flipped tetrahedrons) and does not require polar decomposition.  The energy-based definition of the Neo-Hookean model is as follows:

\begin{equation}
    \Psi_{Neo} = \frac{\lambda}{2} \left(\det (F) - 1\right)^{2} + \frac{\mu}{2} \left(tr(F^{T}F) - 3\right) = \Psi_{H} + \Psi_{D},
\end{equation}
\noindent
where $F$ is a $3 \times 3$ matrix of the deformation gradient and $\lambda$, $\mu$ are the Lamé parameters. $\Psi_{H}$ is the hydrostatic energy resisting expansion and compression. $\Psi_{D}$ is the deviatoric energy resisting distortion. For more details, please refer to \cite{macklin2021constraint}.

In this simulator, we used a constraint-based formulation of the Neo-Hookean energy as recently presented in \cite{macklin2021constraint}. It avoids problems typical for Newton method-based approaches, which require computation of Hessian matrices, their Eigen-decomposition and a sophisticated linear solver. Instead, \cite{macklin2021constraint}  decomposes the energy into hydrostatic (volume preserving) and deviatoric (distortional) constraints. 

This approach applies XPBD's iterative constrained optimization methods, which require only first-order gradients.  Constraints are defined by function $C(x)$ and can be turned into energy using:

\begin{equation}
    \Psi_{C} = \frac{1}{2} \alpha^{-1}C(x)^{2}
\end{equation}
The force $f_{C}$ is then:

\begin{equation}
    f_{C} = -\alpha^{-1}C(x)\frac{\partial C(x)^{T}}{\partial x},
\end{equation}
where $\alpha$ is the compliance (inverse stiffness) of the constraint. In XPBD, the constraints are enforced via Lagrange multipliers $\lambda_{elastic}$:

\begin{equation}
    \lambda_{elastic} = -\overline{\alpha}^{-1}C(x),
\end{equation}
where $\overline{\alpha} = {\alpha}/{\Delta t^{2}}$, where ${\Delta t}$ is simulation time-step.

Taking all the above into consideration, the constraints equations in the XPBD physics simulation framework for the Neo-Hookean Hydrostatic and Deviatoric constraints are as follows:

\begin{equation}
    C_{H}(F) = \det(F) - 1
\end{equation}

\begin{equation}
    C_{D}(F) = \sqrt{tr\left(F^{T}F\right)}
\end{equation}

The compliance parameters ${\alpha}$ of both hydrostatic and deviatoric constraints were visually tuned to approximate the behaviour of the real anatomy based on the judgement of senior clinicians interacting with the virtual anatomy through the laparoscopic device, whilst achieving real-time performance at high physics refresh rate.

\subsubsection{Collision detection and response}

For broad-phase collision detection, we used a dynamic bounding volumes hierarchy of axis-aligned bounding boxes. For the narrow-phase between the virtual tissue and fetoscopic instruments, Signed Distance Fields (SDFs) were applied. Due to their computational performance and robustness, SDFs are popular shape representations for collision detection. It has been well established in the literature how to check individual points against intersections with SDFs. However, until \cite{SDF}, it was unclear how to test continuous surfaces such as triangular and tetrahedral meshes for penetration with SDF. 
We followed  Macklin \cite{SDF} who proposed a per-element local optimization scheme to find the closest points between SDF’s iso-surface and convex elements such as points, edges, triangles and tetrahedrons. This enabled an accurate collision handling between tissues and sharp and pointy surgical tools.

For each intersecting element-SDF pair, an XPBD collision constraint was generated. It pushed the tissue elements out of the collisions with the instrument’s SDF according to penetration depth and normal vectors. We used a Colombian friction approximation, taking into consideration positions from the previous and the current physics steps to approximate the velocities needed for friction calculations \cite{RigidBodies} .

\section{Experiments and Results}
\label{sec:experiments}

We performed face validation of the simulation of the SB repair procedure described in section~\ref{sec:usability}. Fourteen physicians participated and the validation was done using the questionnaire described in Section~\ref{sec:quest}. We present the results in Section~\ref{sec:results}

\subsection{VR SB Repair Simulation Procedure}
\label{sec:usability}

The virtual procedure started with a simulated view of a virtual Spina Bifida (Fig. \ref{fig:vr_surgery}). The participants were required to partially perform three SB-repair procedures, one at a time. Prior to performing their first  procedure, participants’ demographic data (age and gender), operative experience, and interest in virtual reality simulation were obtained through an online questionnaire. Next, all participants were provided with technical instructions outlining the simulation system. The aim  was to give a brief overview of the equipment, tasks and factors, which will differ from real life owing to the limitations of performing the procedure in a simulated setting. The instructions did not instruct on the particular challenges of performing an SB repair. Next, participants were allowed a maximum of 2 minutes to familiarize themselves with how to operate the instruments. Prior to commencing their first procedure, participants were given the opportunity to ask questions relating to the practicalities of the simulation but were not allowed to request any technical advice as to how best to perform the procedure. No time limit was set for the procedures.

\subsection{Questionnaire}
\label{sec:quest}

Face and content validity were evaluated by asking participants to complete the aforementioned questionnaire after finishing their last procedure. The questionnaire assessed the behaviour of tools and tissues, graphical appearance, and overall realism (face validity), as well as the perceived utility of the simulator as a training tool for fetoscopic SB-repair (content validity). The specific questions were as follows:

\begin{enumerate}[label=(\Alph*)]
    \item Rate the realism of the behaviour of virtual tissue,
    \item Rate the realism of the behaviour of virtual laparoscopes,
    \item Rate the realism of the interactions between virtual laparoscopes and tissues,
    \item Rate the realism of the laparoscopic hardware (how realistic it feels),
    \item Rate the visual realism of the simulator (how realistic it looks),
    \item Rate the overall realism of the simulator.
\end{enumerate}

In the questionnaire we used a 5-point Likert scale where 1 corresponded to \textit{very unrealistic/not at all useful} and 5 to \textit{very realistic/very useful}. The average score was calculated by multiplying these scale values by the number of participants who chose the corresponding value and dividing the result by the total number of participants, the standard deviation was calculated by treating the responses as the ordinal data.

\begin{figure*}[t!]
    \centering
    \vspace*{0.2cm}
    \includegraphics[width=16cm, height=8cm]{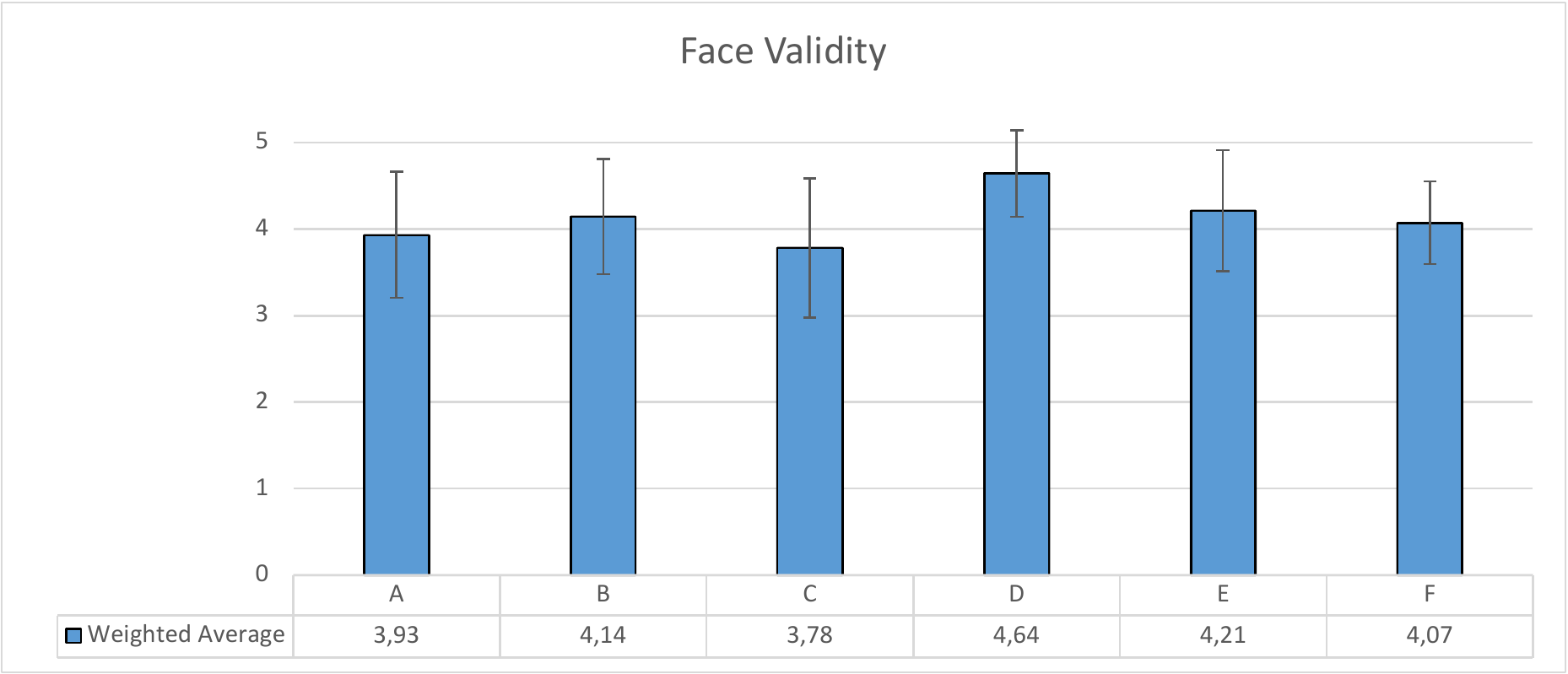}
    \caption{Face validation results. (A) Rate the realism of the behaviour of virtual tissue, (B) Rate the realism of the behaviour of virtual laparoscopes, (C) Rate the realism of the interactions between virtual laparoscopes and tissues, (D) Rate the realism of the laparoscopic hardware (how realistic it feels), (E) Rate the visual realism of the simulator (how realistic it looks), (F) Rate the overall realism of the simulator.}
    \label{fig:face}
\end{figure*}

\subsection{Face validation results}
\label{sec:results}

In total, 14 participants took part in the study (4 male and 10 female). Eight classified their speciality as \textit{Gynecology} and four as \textit{Neonatology}. Two of the participants stated their speciality as Gynecologic Surgery. The average age was 42 years old (SD = 12). Half of the participants (7) had previously used a laparoscopic simulator. Most of the participants stated (85.7\% agreed or strongly agreed) that they are interested in the concept of VR simulation in surgery and 71.4\% that they are interested in simulation in fetoscopic surgery. Four participants (28.6\%) stated that they are not interested in fetoscopic surgery, 3 were neutral and the rest expressed their interest (4) or strong interest (3) in such surgery.

Regarding face validity summarized in Fig.~\ref{fig:face},  the overall simulation realism average score was 4.07 (SD = 0.47) on a 5-point Likert scale described in Section~\ref{sec:quest}.
The participants were most satisfied with the laparoscopic hardware interface (average score 4.64, SD = 0.5), whereas, the realism of virtual laparoscopes scored on average 4.14 (SD = 0.66). The average  score of visual realism was 4.21 (SD = 0.7). The participants were most critical about the interactions between virtual laparoscopes and virtual tissue was 3.78 (SD = 0.8) and the behaviour of virtual tissue, which was marked on average at 3.93 (SD = 0.73).  

The simulator showed very good content validity. Its usefulness as a training tool for SB-repair as well as in learning fundamental laparoscopic skills was marked 4.63 (SD = 0.6) and 4.8 (SD = 0.44), respectively. Finally, the vast majority of the participants (85.7\%) would recommend the simulator to others, whilst 78.6\% of participants found SB-repair simulation as a useful tool in surgical training curricula.

\section{Discussion and Conclusions}
\label{sec:conclusions}

We have developed a prototype VR simulator for core manual skills training for fetoscopic Spina Bifida repair procedures. The simulator currently consists of a circumferential thermocoagulation in the SB tissue followed by its resection using scissors and a grasper. Our goal is to enable surgical trainees to gain their operative skills in a safe environment and provide them with objective feedback on their performance before they consolidate them on patients. 

The initial validation study rated the simulator as realistic, recognising its usefulness as a training tool for fetoscopic procedures and learning fundamental laparoscopic skills. This indicates a potential for further development and possible adoption in resident training. 

Virtual Reality simulation of fetoscopic procedures can contribute to training and improving the educational experience of surgery residents without putting fetuses and their mothers at risk. It could also facilitate wider adaptation of fetoscopic procedures in place of much more invasive open fetal surgeries. However, for an improved educational outcome, such virtual training should be still supplemented by practice on laparoscopic box trainers and/or animal models.


Furthermore, we believe that realistic simulation of surgical procedures will be indispensable for the development of future autonomous surgical robots. Methods such as Reinforcement Learning (RL) can be used to train fully- or semi-autonomous surgical robots in virtual environments orders of magnitude faster and for a fraction of the cost than in the real world. RL is currently widely used in robotic research to solve a variety of complex problems \cite{rl_intro}. However, the vast majority of this work pertains to robots interacting with rigid bodies. At the moment, the research in the application of RL to manipulate soft-bodies is scarce and even scarcer in surgical applications \cite{richter2019open,barnoy2021robotic,tagliabue2020soft}. 
SB repair poses an interesting case for RL. Although it concerns only a single tissue in a relatively small operative field, it requires many fundamental and ubiquitous surgical actions such as tissue grasping and pulling with a grasper, circumferential thermo-coagulation, resection with scissors, irrigation, and suturing. Moreover, these tasks are sequential, one after another, which makes it easier to define rewards and penalties for RL. In future work, we plan to further explore the application of RL in our simulator.

Our work describes a prototype and, as such, it has limitations. The study participants in their free-comments reported realism flaws in collision response between tissues and laparoscopes as well as in haptic feedback. Therefore, we plan to replace our current hardware setup based on generic 3-DoF Geomagic Touch haptic devices with a high-fidelity laparoscopic interface such as Haptic Avatar (Follou AB, Sweden), with 4-DoF haptics. We hope that such an upgrade alone will considerably improve the realism of our simulator. We also plan to further improve the realism of tissue simulation. Finally, we would like to add irrigation and patch suturing tasks to offer a complete SB-repair experience. Larger usability studies, such as construct validity and transfer of skills, are envisaged as the prototype is further developed. 

\section{Acknowledgements}
The authors wish to thank the study participants. This work is supported in part by the European Union’s Horizon 2020 research and innovation programme under grant agreement Sano No. 857533 and the International Research Agendas programme of the Foundation for Polish Science, co-financed by the EU under the European Regional Development Fund. 

\bibliographystyle{IEEEtran}

\bibliography{IEEEabrv,references}

\end{document}